\documentclass[10pt,conference]{IEEEtran}

\newif\ifieee
\ieeefalse

\usepackage[sort,compress]{cite}

\ifCLASSINFOpdf
   \usepackage[pdftex]{graphicx}
   \graphicspath{{figs/}}
   \DeclareGraphicsExtensions{.pdf,.jpeg,.png}
\else
   \usepackage[dvips]{graphicx}
   \graphicspath{{../figs/}}
   \DeclareGraphicsExtensions{.eps}
\fi

\usepackage[cmex10]{amsmath}
\usepackage[dvipsnames,table,xcdraw]{xcolor}
\usepackage{amsthm}

\usepackage{algorithmicx}
\usepackage[ruled]{algorithm}
\usepackage{algpseudocode}

\usepackage{array}

\ifCLASSOPTIONcompsoc
  \usepackage[caption=false,font=normalsize,labelfont=sf,textfont=sf,farskip=0pt]{subfig}
\else
  \usepackage[caption=false,font=footnotesize,farskip=0pt]{subfig}
\fi

\usepackage[hyphens]{url}

\usepackage{xcolor}
\usepackage{amsfonts} 
\usepackage{booktabs}

\newif\iffinal
\finaltrue
\newcommand{\cmtid}{72}
 \newcommand{\papertitle}{\huge Face Super-Resolution Using Stochastic Differential Equations} %

\iffinal
\else
\usepackage[switch]{lineno}
\fi

\usepackage[normalem]{ulem}

\newcommand\MS[1]{{#1}} %

\newcommand*{\DM}[2][]{\textcolor{ForestGreen}{[\textbf{\ifthenelse{\equal{#1}{}}{DM}{DM(#1)}}: #2]}}
\newcommand*{\RL}[2][]{\textcolor{violet}{[\textbf{\ifthenelse{\equal{#1}{}}{RL}{RL(#1)}}: #2]}}
\newcommand*{\JN}[2][]{\textcolor{blue}{[\textbf{\ifthenelse{\equal{#1}{}}{JN}{JN(#1)}}: #2]}}
\newcommand*{\HP}[2][]{\textcolor{red}{[\textbf{\ifthenelse{\equal{#1}{}}{HP}{HP(#1)}}: #2]}}

\ifieee
\else
\usepackage[colorlinks=true,allcolors=black]{hyperref}
\fi
\usepackage[acronym]{glossaries}
\usepackage{xspace}
\usepackage[capitalise]{cleveref} %
\usepackage[T1]{fontenc}
\usepackage{balance}

\newcommand\red[1]{{\textcolor{red}{#1}}}
\newcommand\blue[1]{{\textcolor{blue}{#1}}}

\newacronymstyle{long-short-br}
{%
  \GlsUseAcrEntryDispStyle{long-short}%
}%
{%
  \GlsUseAcrStyleDefs{long-short}%
}
\setacronymstyle{long-short-br}

\begin{document}
\title{\huge\papertitle}

\iffinal

\author{Marcelo dos Santos\IEEEauthorrefmark{1}, Rayson Laroca\IEEEauthorrefmark{1}, Rafael O. Ribeiro\IEEEauthorrefmark{2}, Jo\~{a}o Neves\IEEEauthorrefmark{3}, Hugo Proen\c{c}a\IEEEauthorrefmark{3}, David Menotti\IEEEauthorrefmark{1}\\
\IEEEauthorrefmark{1}\hspace{0.2mm}Department of Informatics, Federal University of Paran\'a, Curitiba, Brazil\\
\IEEEauthorrefmark{2}\hspace{0.2mm}National Institute of Criminalistics, Brazilian Federal Police, Bras\'{\i}lia, Brazil\\
\IEEEauthorrefmark{3}\hspace{0.2mm}Instituto de Telecomunica\c{c}\~{o}es, University of Beira Interior, Covilh\~{a}, Portugal\\
\resizebox{0.95\linewidth}{!}{
\IEEEauthorrefmark{1}{\tt\small \{msantos,rblsantos,menotti\}@inf.ufpr.br} \quad
\IEEEauthorrefmark{2}{\tt\small rafael.ror@pf.gov.br} \quad
\IEEEauthorrefmark{3}{\tt\small \{jcneves,hugomcp\}@di.ubi.pt}
}
}

\else
  \author{SIBGRAPI paper ID: \cmtid \\[10ex] }
  \linenumbers
\fi

\maketitle

\ifieee
{\let\thefootnote\relax\footnote{\\978-1-6654-5385-1/22/\$31.00\textcopyright2022 IEEE}}
\else
\fi

\newacronym{ddpm}{DDPM}{Denoising Diffusion Probabilistic Models}
\newacronym{ffhq}{FFHQ}{Flickr-Faces-HQ}
\newacronym{gan}{GAN}{Generative Adversarial Network}
\newacronym{hr}{HR}{high-resolution}
\newacronym{mcmc}{MCMC}{Markov chain Monte Carlo}
\newacronym{mse}{MSE}{Mean Squared Error}
\newacronym{psnr}{PSNR}{peak signal-to-noise ratio}
\newacronym{smld}{SMLD}{Score matching with Langevin dynamics}
\newacronym{sr}{SR}{super-resolution}
\newacronym{lr}{LR}{low-resolution}
\newacronym{ssim}{SSIM}{structural similarity index measure}
\newacronym{sde}{SDE}{stochastic differential equation}
\newacronym{ve}{VE}{variation exploding}
\newacronym{vp}{VP}{variation preserving}

\newcommand{\ffhq}{\gls*{ffhq}\xspace}
\newcommand{\gfpgan}{GFP-GAN\xspace}
\newcommand{\sparnet}{SPARNet\xspace}

\iffinal
\newcommand{\supplementary}{\url{https://github.com/marcelowds/sr-sde}}

\else

\newcommand{\supplementary}{[hidden for review]}

\fi

\ifieee
\vspace{-3.7mm}
\else
\fi
\begin{abstract}
Diffusion models have proven effective for various applications such as images, audio and graph generation.
Other important applications are image super-resolution and the solution of inverse problems.
More recently, some works have used \glspl*{sde} to generalize diffusion models to continuous time.
In this work, we introduce \glspl*{sde} to generate super-resolution face images.
To the best of our knowledge, this is the first time \glspl*{sde} have been used for such an application.
The proposed method provides an improved \gls*{psnr}, \gls*{ssim}, and consistency than the existing super-resolution methods based on diffusion models. 
In particular, we also assess the potential application of this method for the face recognition task.
A generic facial feature extractor is used to compare the super-resolution images with the ground truth, and superior results were obtained compared with other~methods.
Our code is publicly available at \textit{\supplementary}.
\end{abstract}

\IEEEpeerreviewmaketitle

\section{Introduction}
\label{sec:introduction}

\glsresetall

Probabilistic approaches have been successfully applied to data generation.
Two main discrete models are known as \gls*{ddpm}~\cite{sohl2015deep, ho2020denoising} and \gls*{smld}~\cite{song2019generative}.

Inspired by considerations from non-equilibrium thermodynamics, in the \gls*{ddpm}, a Markov chain is used to model the forward and reverse processes of a diffusion model.
In the forward process, random Gaussian noise is added to the clean image until a pure noisy image is obtained.
A network is trained to predict the noise level of the image at each step.
To generate an image in the reverse process, a pure Gaussian noise is considered as the initial state of a Markov chain.
The network is used to iteratively denoise the image until a clean image is obtained.

Similar to \gls*{ddpm}, the \gls*{smld} model consists of perturbing the data with different scales of random Gaussian noise.
A network conditioned at the noise level is trained to learn the gradient of the log probability density with respect to data.
The Langevin dynamics~\cite{sarkka2019applied} is used for the data generation process to remove the noise from the data iteratively.
Starting from high noise levels, the process runs until low noise levels are reached and the generated images have distributions indistinguishable from the original data distribution.
These two classes of models are part of score-based generative models. 

Score-based generative models have been successfully applied to a wide range of different data, such as the generation of audio~\cite{chen2020wavegrad}, graphs~\cite{niu2020permutation} and shapes~\cite{cai2020learning} as well as for image synthesis, achieving results even better than \glspl*{gan}~\cite{dhariwal2021diffusion}, image edition~\cite{meng2022sdedit}, text-to-image translation~\cite{saharia2022photorealistic}, general inverse problems~\cite{song2022solving, kawar2021snips}, super-resolution~\cite{saharia2021image, li2022srdiff}, among others. 
For the super-resolution task, Saharia et al.~\cite{saharia2021image} and Li et al.~\cite{li2022srdiff} adapted the \gls*{ddpm} models to generate super-resolution images using low-resolution images as a guide to a  Markov chain. 
In~\cite{kawar2021snips}, super-resolution was treated as a special case of an inverse~problem.

In~\cite{song2021score}, the authors presented a generalization of score-based models to continuous time using a \gls*{sde}.
The \gls*{ddpm} and \gls*{smld} models are considered particular cases of a general \gls*{sde}. 
The generalization of the \gls*{ddpm} and \gls*{smld} models are called \gls*{vp} and \gls*{ve}, respectively.
In~\cite{song2021score}, it is also presented a third model named subVP, which has the variance preserved during the diffusion process but is limited by the variance of the \gls*{vp} process. 
The developed methods reached record-breaking results for unconditional image generation, which is performed by solving a reverse \gls*{sde}.
More recently, in~\cite{jolicoeur2021gotta,vahdat2021score,dockhorn2022score}, the \gls*{sde} score-based models are further developed, focusing on decreasing the execution time, optimizing the image generation process, and improving high-resolution image~synthesis.

In this work, the continuous version of diffusion models described through an \gls*{sde} in \cite{song2021score} is adapted to deal with the face super-resolution problem.
As far as we know, this is the first time that \glspl*{sde} are used to  face super-resolution images.
We evaluate four different algorithms: \gls*{sde}-\gls*{vp}, \gls*{sde}-subVP, \gls*{sde}-\gls*{ve} and \gls*{sde}-\gls*{ve}cs.
The robustness of the obtained models is demonstrated by the generation of detailed and high-quality super-resolution~images. 

The \gls*{psnr} and \gls*{ssim} values achieved are similar to those obtained by other super-resolution methods based on diffusion models.
However, considering that these metrics are not strongly correlated with how humans perceive image quality~\cite{zhang2018unreasonable, saharia2021image}, we conducted an additional experiment to demonstrate the superiority of the proposed method.
The \gls*{hr} and \gls*{sr} images are provided to a VGG-Face network~\cite{vggface} for obtaining a compact feature descriptor and measuring the average cosine similarity between the corresponding HR and SR images. The obtained results are more promising than existing methods. We also show that if we only rely on the \gls*{psnr} metric, this does not always provide the best images for face recognition. 
Moreover, our super-resolution method is based on denoising, so our method is particularly suitable for surveillance scenarios where the image quality is low and~noisy.

The remaining of this paper is organized as follows: Section~\ref{sec:Theoretical} contains the theoretical background behind the proposed method, while Section~\ref{sec:Exp_and_Res} describes our experiments and results. Finally, section~\ref{sec:Conclusions} concludes the paper.

\section{Theoretical background}
\label{sec:Theoretical}
\subsection{Stochastic differential equations and super-resolution}

A continuous diffusion process $\{x(t)\}_{t=0}^T$ can be modeled by the Itô \gls*{sde} 
\begin{equation}
    \mathrm{d}\mathbf{x}=\mathbf{f}(\mathbf{x},t)\mathrm{d}t+g(t)\mathrm{d}\mathbf{w},
    \label{eqsde}
\end{equation}
\noindent where $\mathbf{f}(\mathbf{x},t)$ is the drift coefficient, $g(t)$ is a diffusion coefficient, and $\mathbf{w}$ is a Wiener process. For more details about Itô \gls*{sde} and Wiener process, see~\cite{sdeplaten, sarkka2019applied}. In~\cite{anderson1982reverse}, it was shown that it is possible to reverse the diffusion process (Eq.~\ref{eqsde}) using another diffusion process given by
\begin{equation}
    \mathrm{d}\mathbf{x}=[\mathbf{f}(\mathbf{x},t)-g(t)^2\nabla_{\mathbf{x}}\log p_t(\mathbf{x})]\mathrm{d}t+g(t)\mathrm{d}\bar{\mathbf{w}},
    \label{revsde}
\end{equation}
\noindent where $\mathrm{d}\bar{\mathbf{w}}$ is a Wiener process running backwards in time.

\begin{figure}[!htb]
\centering
\includegraphics[width=0.8\linewidth]{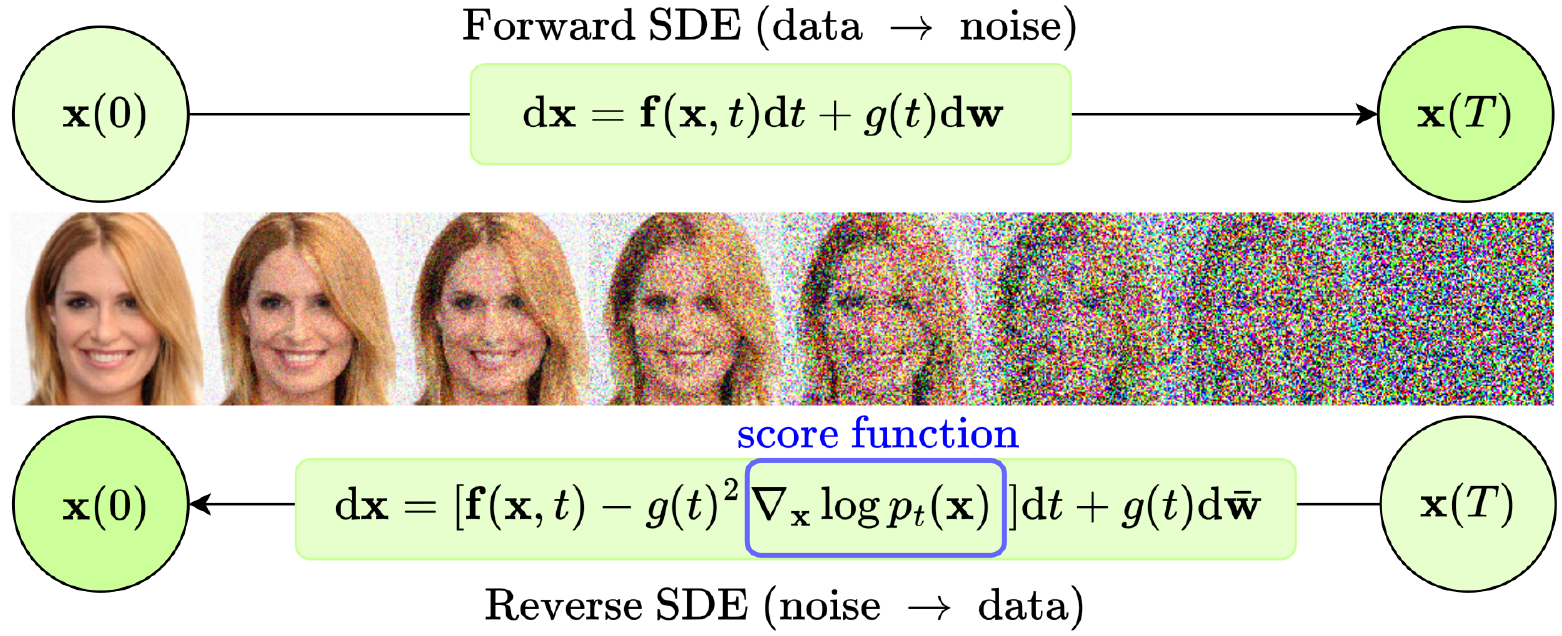}

\vspace{-3mm}

\caption{Forward and reverse processes. Adapted from~\cite{song2021score}.}
\label{fig_sim}
\end{figure}

In~\cite{song2021score}, the authors used a deep neural network $s_{\theta}(\mathbf{x}(t),t)$ to learn $\nabla_{\mathbf{x}}\log p_t(\mathbf{x})$ during the forward process.
Once this gradient  is learned,  the reverse process (Eq.~\ref{revsde}) is run, changing $\nabla_{\mathbf{x}}\log p_t(\mathbf{x})$  by $s_{\theta}(\mathbf{x}(t),t)$  to generate a sample $\mathbf{x}(0)\sim p_0$ (see \cref{fig_sim}).
As we are dealing with the super-resolution problem, in this work, we also use a down-sampled version $\mathbf{y}$ of the image $\mathbf{x}$ as input to the network, denoted by $s_{\theta}(\mathbf{x}(t),\mathbf{y},t)$.
To generate a super-resolution image, during the reverse process we initially use a noisy image $\mathbf{x}(T)$ to iteratively transform it on $\mathbf{x}(0)$, using $\mathbf{y}$ as a  guide to the stochastic process.

To train the network, we must find the parameters $\theta$ of  $s_{\theta}(\mathbf{x}(t),\mathbf{y},t)$ such that $\nabla_{\mathbf{x}}\log p_t(\mathbf{x})$ can be well approximated by the network. For this purpose, we must optimize the following function~\cite{vincent2011connection}
\begin{align}
     \min_{\theta}
    \mathbb{E}_{t\sim \mathcal{U}[0,T]}
   \mathbb{E}_{\mathbf{x}_0\sim p(\mathbf{x}_0)}\mathbb{E}_{\mathbf{x}(t)\sim p_t(\mathbf{x}(t)|\mathbf{x}(0)}
    \big[\lambda(t)\nonumber\\ \times\, \|s_\theta(\mathbf{x}(t),\mathbf{y},t)-\nabla_{\mathbf{x}(t)}\log p_{}(\mathbf{x}(t)|\mathbf{x}(0))\|_2^2\big],
    \label{eqloss}
\end{align}
where $\lambda(t)$ is a positive weighting function. 
In Eq.~\ref{eqloss}, we need to compute $p(\mathbf{x}(t)|\mathbf{x}(0))$, but if $\mathbf{f}(\mathbf{x},t)$  from Eq.~\ref{eqsde} is an affine function,  the distribution $p(\mathbf{x}_t|\mathbf{x}_0)$ is a normal distribution where the mean $\pmb{\mu}(t)$ and variance $\pmb{\Sigma}(t)$ evolve according to the following  differential equations~\cite{sarkka2019applied}
\begin{equation}
    \frac{\mathrm{d}\pmb{\mu}(t)}{\mathrm{d}t}=\mathbb{E}[\mathbf{f}(\mathbf{x},t)]
    \label{eqmu}
\end{equation}
and 
\begin{align}
    \frac{\mathrm{d}\pmb{\Sigma}(t)}{\mathrm{d}t}=\mathbb{E}[\mathbf{f}(\mathbf{x},t)(\mathbf{x}-\pmb{\mu}(t))^\top+
    (\mathbf{x}-\pmb{\mu}(t))\mathbf{f}^\top(\mathbf{x},t)]\nonumber\\+g^2(t)\mathbf{I}.
    \label{eqsigma}
\end{align}

In the \gls*{ve}, \gls*{vp} and subVP cases described in~\cite{song2021score} we have $\mathbf{f}(\mathbf{x},t)$ and $g(t)$  given respectively by
\begin{equation}
\mathbf{f}(\mathbf{x},t)=\mathbf{0},\quad g(t)=\sqrt{\frac{\mathrm{d}\sigma^2(t)}{\mathrm{d}t}},    
\end{equation}
\begin{equation}
\mathbf{f}(\mathbf{x},t)=-\frac{1}{2}\beta(t)\mathbf{x},\quad g(t)=\sqrt{\beta(t)}    
\end{equation}
and 
\begin{equation}
\mathbf{f}(\mathbf{x},t)=-\frac{1}{2}\beta(t)\mathbf{x},\, g(t)=\sqrt{\beta(t)\left(1-e^{-2\int_0^t\beta(s)\mathrm{d}s}\right)},   
\end{equation}
\noindent where $\sigma(t)$ and $\beta(t)$ are functions which describe the level of noise added to the data at each time. 

For all the three models, the drift coefficients are affine functions and the mean and variance are computed analytically using  Eq.~\ref{eqmu} and Eq.~\ref{eqsigma}.
The results for the mean and variance of the  \gls*{ve}, \gls*{vp} and subVP models are obtained in~\cite{song2021score} and are given respectively by
\begin{equation}
\pmb{\mu}(t) = \mathbf{x}(0), \,\, \pmb{\Sigma}(t)=[\sigma^2(t)-\sigma^2(0)]\mathbf{I},    
\end{equation}
\begin{equation}
\pmb{\mu}(t) =\mathbf{x}(0)e^{-\frac{1}{2}\int_0^t\beta(s)\mathrm{d}s}, \,\, \pmb{\Sigma}(t)=[1-e^{-\frac{1}{2}\int_0^t\beta(s)\mathrm{d}s}]\mathbf{I},    
\end{equation}
and
\begin{equation}
\pmb{\mu}(t) =\mathbf{x}(0)e^{-\frac{1}{2}\int_0^t\beta(s)\mathrm{d}s}, \,\, \pmb{\Sigma}(t)=[1-e^{-\frac{1}{2}\int_0^t\beta(s)\mathrm{d}s}]^2\mathbf{I}.    
\end{equation}
With these three equations, it is possible to compute $p(\mathbf{x}_t|\mathbf{x}_0)$ and  calculate   the loss function (Eq.~\ref{eqloss}) during the training~process.

Following~\cite{song2021score}, we choose  $\sigma(t)=\sigma_{min}\left(\sigma_{max}/\sigma_{min}\right)^t$ and $\beta(t)=\bar{\beta}_{min}+(\bar{\beta}_{max}-\bar{\beta}_{min})t$ to describe the noise level.

\subsection{Solving the SDE}

After the training is performed and the function $s_{\theta}$ is obtained, we need to solve the reverse diffusion (Eq.~\ref{revsde}) from $t=T$ to $t=0$ in order to obtain the super-resolution images.
For this purpose, we use the Algorithm (\ref{alg1}), which is composed of two main functions, the Predictor and the Corrector. 

The Predictor function gives an estimate of the sample at the next time step. 
For the \gls*{sde}-\gls*{ve} model it is used the Euler-Maruyama~\cite{sdeplaten, sarkka2019applied}, while for the other models, it is used the \emph{Reverse-diffusion} discretization strategy~\cite{song2021score}, which is a simple discretization of the reverse \gls*{sde}, given by Eq.~\ref{revsde}.

The Corrector function corrects the marginal distribution of the estimated sample \cite{song2021score}.
This is done by combining numerical \gls*{sde} solvers with score-based \gls*{mcmc} approaches, such as Langevin Monte Carlo Markov Chain~\cite{parisi1981correlation} or Hamiltonian Monte Carlo method~\cite{neal2011mcmc}.
For more details about Predictor-Corrector steps, see~\cite{song2021score}. 

In this work, the models \gls*{sde}-\gls*{vp} and \gls*{sde}-subVP use no corrector method, i.e., the Corrector is equal to identity.
The \gls*{sde}-\gls*{ve} method is evaluated in two versions: one with the identity corrector step and another with the Langevin corrector step, which we call \gls*{sde}-\gls*{ve}cs.

\begin{algorithm}[!ht]
\caption{Predictor-Corrector (PC) sampling}\label{alg1}
N: Number of discretization steps for the reverse-time \gls*{sde}\\
M: Number of correction steps
\begin{algorithmic}[1]
\State Initialize $\mathbf{x}_T \sim p_T(\mathbf{x})$
\For {$i=N-1$ \textbf{to} $0$}
\State $\mathbf{x}_i\gets \text{Predictor} (\mathbf{x}_{i+1})$
\For {$j=1$ \textbf{to} $M$}
\State $\mathbf{x}_i\gets \text{Corrector} (\mathbf{x}_{i})$
\EndFor
\EndFor
\State \textbf{return} $\mathbf{x}_0$%
\end{algorithmic}
\end{algorithm}

\subsection{Network architecture}

\begin{figure}[!ht]
\centering
\includegraphics[width=0.875\linewidth]{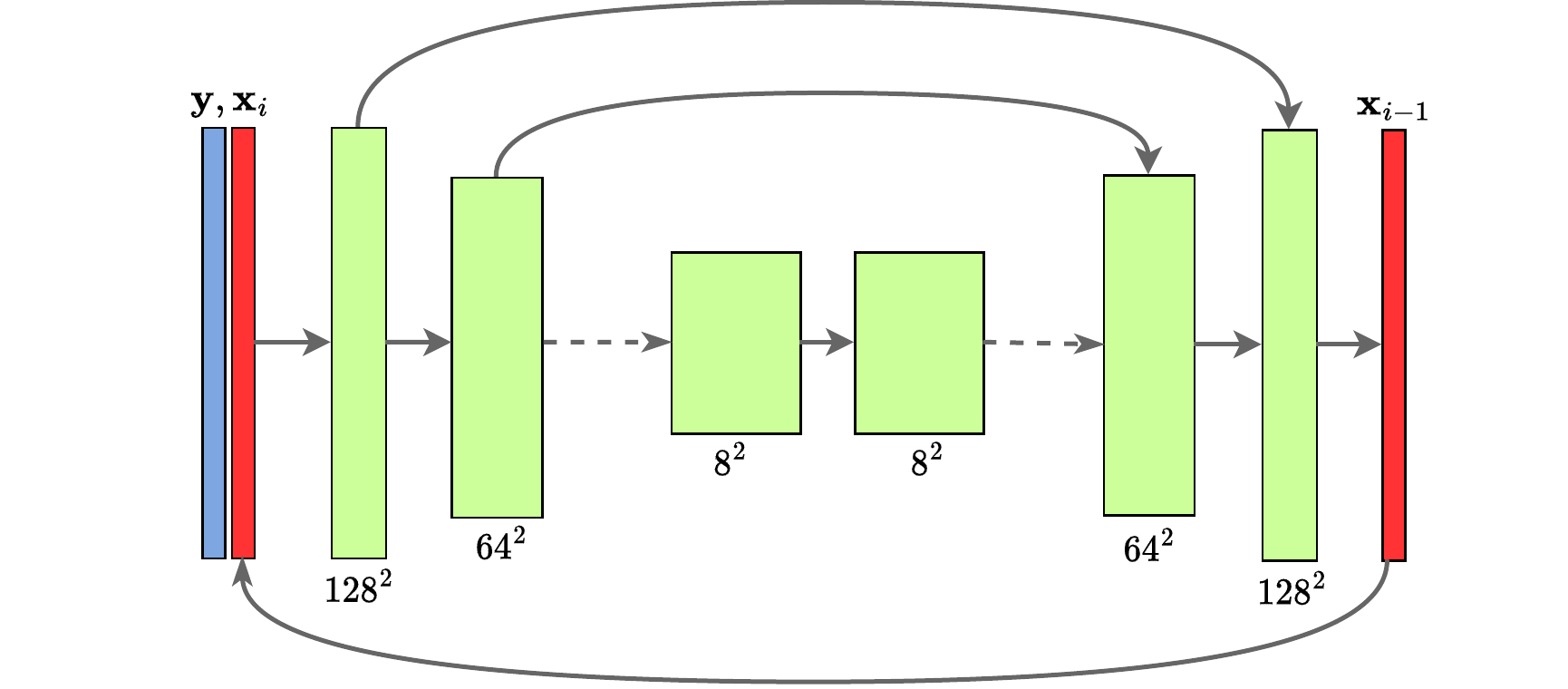}

\vspace{-2mm}

\caption{Architecture of the network.
The concatenation of the low-resolution image $\mathbf{y}$ and the noisy image $\mathbf{x}_i$ is used as input to the network. Adapted from~\cite{saharia2021image}.}
\label{fig_unet}
\end{figure}

The network architecture we use is mainly inspired by the U-net architecture~\cite{ho2020denoising} with the improvements described in~\cite{song2021score} but adapted to receive the \gls*{lr} image, similar to the adaptation made by Saharia et al.~\cite{saharia2021image}.
The \gls*{lr} image is upsampled to the target \gls*{hr} size and concatenated with a noisy image.
Therefore, the network receives as input the \gls*{lr} image $\mathbf{y}$ and a noisy image $\mathbf{x}_{i}$, and with the network's output, it is possible to update  $\mathbf{x}_{i}$ to $\mathbf{x}_{i-1}$ using Algorithm (\ref{alg1}) (see \cref{fig_unet}).
This process is repeated from $t=T$ up to $t=0$, where the SR image $\mathbf{x}_{0}$ is obtained.

The source code and the network weights are publicly available at~\textit{\supplementary}.
\section{Experiments and Results}
\label{sec:Exp_and_Res}

Here, we describe the proposed experiments and results obtained with our methods. In  \cref{subseca} we describe the main metrics used to verify the quality of the \gls*{sr} images. In \cref{subsecb} we describe how to use image features and cosine similarity CS to compare \gls*{sr} and \gls*{hr} images. We also analyze how the image smoothness can influence the \gls*{psnr} and CS metrics. %
In \cref{subsecc} we demonstrate the superior quality of our method both qualitatively (with very detailed \gls*{sr} images) and quantitatively with the highest value for the CS metric, which is very important for recognition tasks. Finally, in \cref{subsecd} we describe the experimental parameters used in our experiments.

To evaluate the capacity of \glspl*{sde} for the super-resolution task,  the four algorithms \gls*{sde}-\gls*{vp}, \gls*{sde}-subVP, \gls*{sde}-\gls*{ve} and \gls*{sde}-\gls*{ve}cs  are provided with $16\times 16$ images and produce  $128\times 128$ images. The working size of the network is $128 \times 128$, so the input is upsampled to this size. %

\subsection{Super-resolution  metrics and image smoothness}
\label{subseca}

The metrics \gls*{psnr} and \gls*{ssim} are used to evaluate the quality of the \gls*{sr} images. We also use the consistency, defined as the \gls*{mse} between the down-sampled \gls*{sr} images and the \gls*{lr} inputs.
Aiming to evaluate the potential applications of our method  on face recognition systems, we rely on the cosine similarity between the image features obtained from VGG-Face~\cite{vggface}.

\begin{figure}[!ht]
\centering
\setlength{\tabcolsep}{2pt}

\begin{tabular}{cccc}
\footnotesize{Original} & \footnotesize{Sample1} & \footnotesize{Sample2} & \footnotesize{Sample3} \\
\includegraphics[width=18mm]{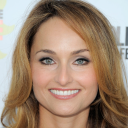} & \includegraphics[width=18mm]{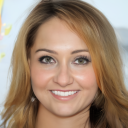} & \includegraphics[width=18mm]{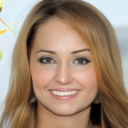} & \includegraphics[width=18mm]{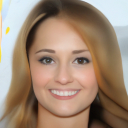} \\
\includegraphics[width=18mm]{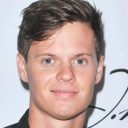} & \includegraphics[width=18mm]{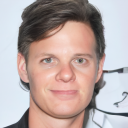} & \includegraphics[width=18mm]{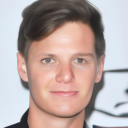} & \includegraphics[width=18mm]{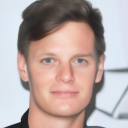}
\end{tabular}

\vspace{-1mm}

\caption{Samples obtained increasing $r$ from left to right. For Sample1, Sample2 and Sample3 the values of $r$ are $0.10$, $0.30$ and $0.52$ for the first row and $0.12$, $0.32$ and $0.39$ for the second row. Higher values of $r$ yield smoother images and larger values of \gls*{psnr} (on~average).}
\label{fig_sr2}
\end{figure}

Specifically for the \gls*{sde}-\gls*{ve}cs method, the Langevin correction depends on a parameter $r$ which controls the signal-to-noise ratio during the correction step.
Higher values of $r$ typically result in smoother images, as can be seen in \cref{fig_sr2}.
It is well known that the \gls*{psnr} and \gls*{ssim} metrics are not adequate for measuring the quality of super-resolution images because these metrics tend to be very conservative with high-frequency details~\cite{johnson2016perceptual,saharia2021image}.
Therefore, we use the \gls*{sde}-\gls*{ve}cs algorithm to generate a set of images with different values of $r$ and analyze the influence of the smoothness level on the \gls*{psnr} metric.

\begin{figure}[!htb]
\centering
\includegraphics[width=0.875\linewidth]{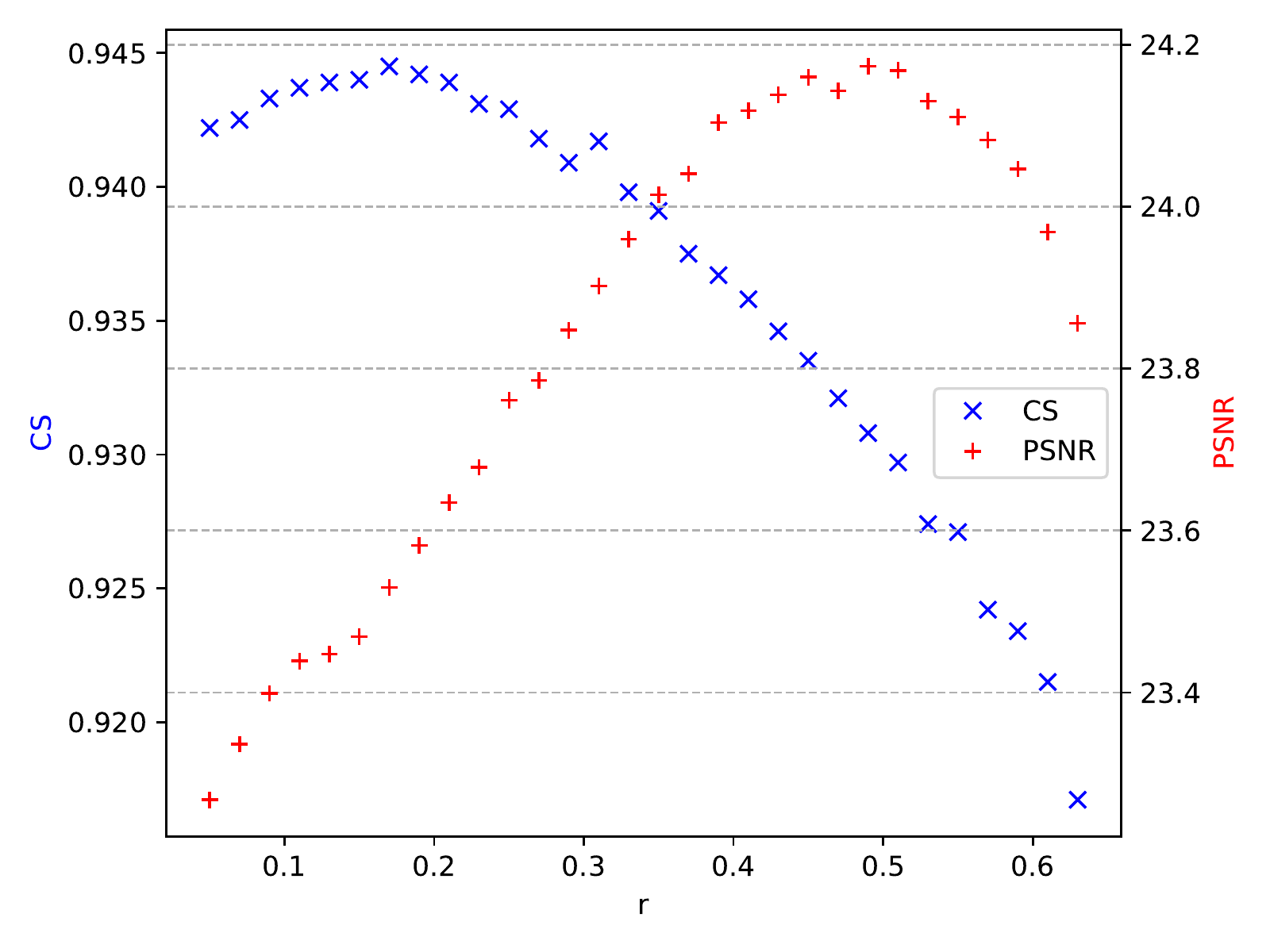}
\vspace{-4.25mm}
\caption{CS and \gls*{psnr} as a function of $r$ (sampling parameter).
Higher values of $r$ produce smoother images (with higher values of \gls*{psnr}) but can decrease the value of~CS.}
\label{fig_psnr}
\end{figure}

Fig.~\ref{fig_psnr} shows \gls*{psnr} as a function of the parameter~$r$.
As can be seen, higher values of $r$ cause the \gls*{psnr} to increase, with the highest value of \gls*{psnr} occurring at $r~\approx~0.5$.
Thus, with smoother images, we can obtain higher values of \gls*{psnr}.
This is in line with the regression model of~\cite{saharia2021image}, which provides blurry and less detailed images than SR3, but with better results for \gls*{psnr} and \gls*{ssim}. 

\subsection{Feature extraction for face recognition}
\label{subsecb}

One application of super-resolution algorithms is face recognition.
In real-world scenarios, the images of surveillance cameras are usually noisy and low resolution~\cite{juefei2019, li2019low, goncalves2019multitask}.
Hence, considering that our method is based on denoising, we believe it can be very suitable for such~scenarios.

To evaluate the potential influence of the smoothness and \gls*{psnr} metric on the recognition accuracy, we use different images obtained with the \gls*{sde}-\gls*{ve}cs algorithm for different values of $r$, analyzing which of these images are better for the face recognition task.
To compare the \gls*{sr} and \gls*{hr} images, we extract the features of the images using VGG-Face and measure the cosine similarity between~them. 

Let $\mathbf{x}$ and $\tilde{\mathbf{x}}$ be the super-resolved and the original images, respectively. 
The features extracted with VGG-Face is a one-dimensional vector of size $F$ and can be denoted by $\mathbf{z}=f(\mathbf{x})$ for the super-resolution images and $\tilde{\mathbf{z}}=f(\tilde{\mathbf{x}})$ for the original images; $f(\cdot)$ is a feature extractor function. 
The cosine similarity between these two vectors is computed using
\begin{equation}
    s(\tilde{\mathbf{z}},\mathbf{z})=\frac{\langle \tilde{\mathbf{z}}, \mathbf{z}\rangle}{\|\tilde{\mathbf{z}}\|\cdot \|\mathbf{z}\|},
\end{equation}
where $\langle\cdot,\cdot\rangle$ denotes the scalar product, and $\|\cdot \|$ refers to the Euclidean norm. 
As the above similarity is calculated for $L$ images, we denote each feature vector by an index $i$. 
The average cosine similarity for the $L$ images, denoted by CS, is 
\begin{equation}
    \text{CS}=\frac{1}{L}\displaystyle\sum_{i=1}^Ls(\tilde{\mathbf{z}}_i,\mathbf{z}_i).
\end{equation}

CS is the result of facial feature extraction, so it strongly influences recognition accuracy.
CS values close to 1 indicate that the features of \gls*{sr} and \gls*{hr} images are very close and that the super-resolution algorithm is retrieving image details important for facial recognition.

\begin{figure}[!ht]
\centering
\includegraphics[width=0.875\linewidth]{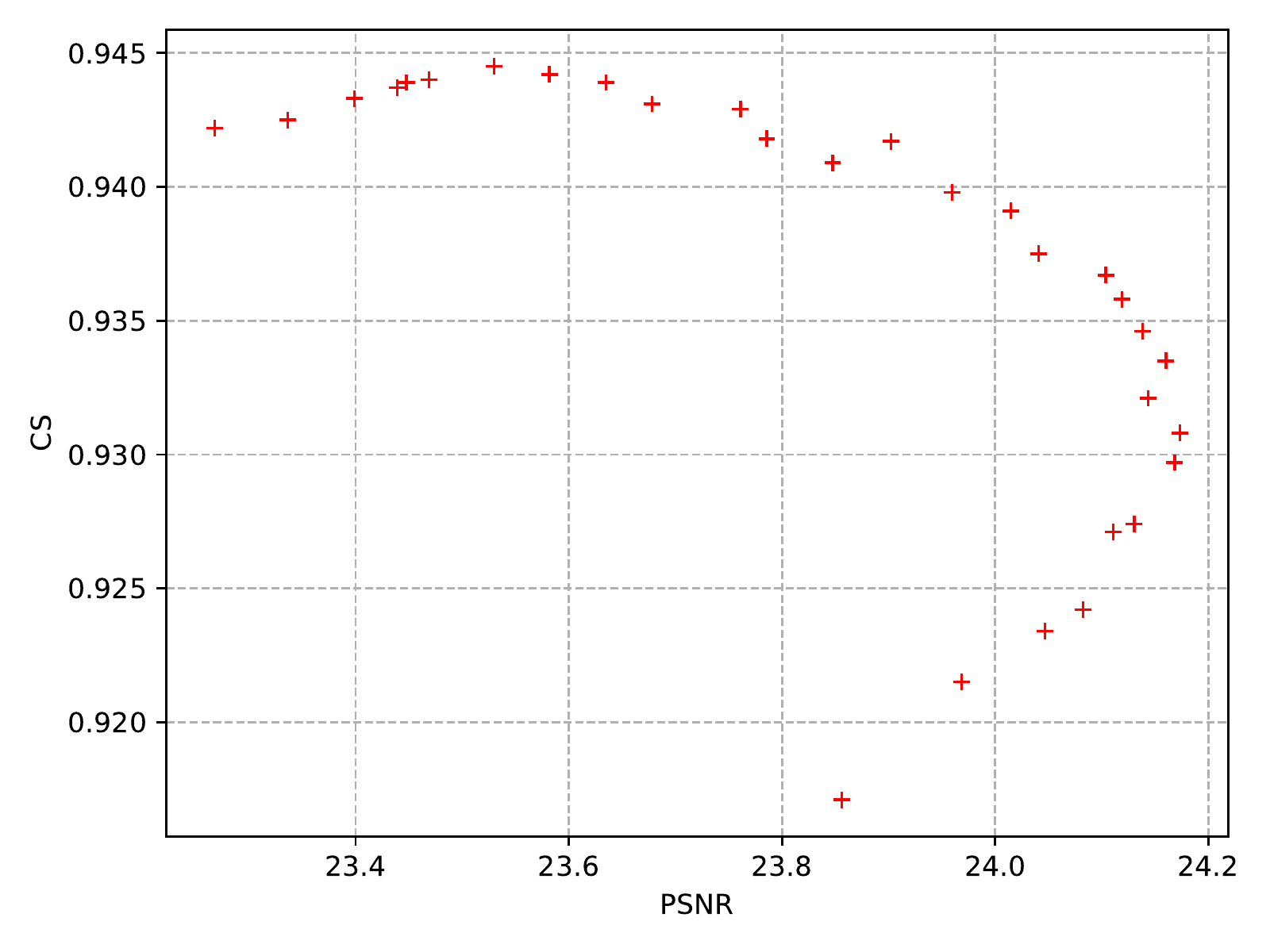}
\vspace{-4mm}
\caption{Cross-plot between CS and \gls*{psnr}. The correlation coefficient between CS and \gls*{psnr}  is  $-0.6591$, implying that higher values of \gls*{psnr} do not always result in higher values of~CS.}
\label{fig_cs_psnr}
\end{figure}

As described in~\cite{johnson2016perceptual,zhang2018unreasonable}, the \gls*{psnr} metric does not align well with human perception. Now one may ask what are the best super-resolution images to use on the recognition tasks, for example, the images with high-frequency details (smaller values of $r$ and \gls*{psnr}) or smoother images (higher values of $r$ and \gls*{psnr}). To answer this question, we analyze CS, as a function of~$r$ in Fig.~\ref{fig_psnr}. We observe that CS has a maximum value for $r~\approx~0.16$ (which is different from~$r$ for which \gls*{psnr} is maximum) and slightly decreases with~$r$. We can also analyze if there is relation between the values of CS and \gls*{psnr}, and for this purpose, we computed the correlation coefficient between CS and \gls*{psnr}, obtaining a value of $-0.6591$. Therefore, not always higher values of \gls*{psnr} are related to higher values of CS (and consequently higher recognition accuracy), as can be observed in the cross-plot between CS and \gls*{psnr} in \cref{fig_cs_psnr}.

\subsection{Super-resolution}
\label{subsecc}
\begin{figure*}[!htb]
\centering
\setlength{\tabcolsep}{2pt}

\resizebox{0.925\linewidth}{!}{ %
\begin{tabular}{ccccccccc}
Original & Input & \small{ \gfpgan~\cite{wang2021gfpgan}\phantom{i}} & \small{\sparnet~\cite{chen2020learning}} & SR3~\cite{saharia2021image} & \textcolor{blue}{SDE-VP} & \textcolor{blue}{SDE-subVP} & \textcolor{red}{SDE-VE} & \textcolor{blue}{SDE-VEcs} \\
\includegraphics[width=22mm]{figs/10-hr.png} & \includegraphics[width=22mm]{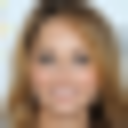} & \includegraphics[width=22mm]{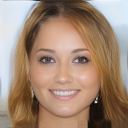} & \includegraphics[width=22mm]{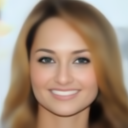} & \includegraphics[width=22mm]{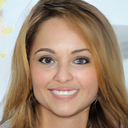} & \includegraphics[width=22mm]{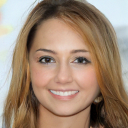} & \includegraphics[width=22mm]{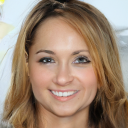} & \includegraphics[width=22mm]{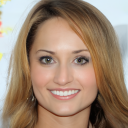} & \includegraphics[width=22mm]{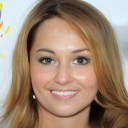} \\
\includegraphics[width=22mm]{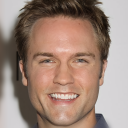} & \includegraphics[width=22mm]{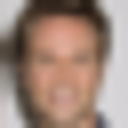} & \includegraphics[width=22mm]{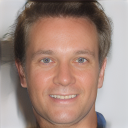} & \includegraphics[width=22mm]{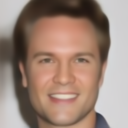} &  \includegraphics[width=22mm]{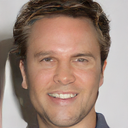} & \includegraphics[width=22mm]{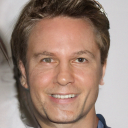} & \includegraphics[width=22mm]{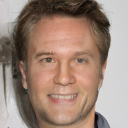} & \includegraphics[width=22mm]{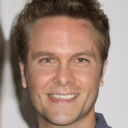} & \includegraphics[width=22mm]{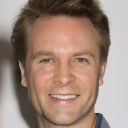} \\
\includegraphics[width=22mm]{figs/6-hr.png} & \includegraphics[width=22mm]{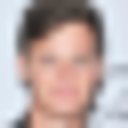} & \includegraphics[width=22mm]{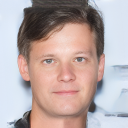} & \includegraphics[width=22mm]{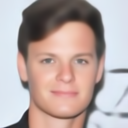} & \includegraphics[width=22mm]{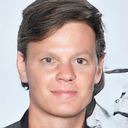} & \includegraphics[width=22mm]{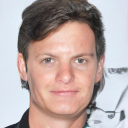} & \includegraphics[width=22mm]{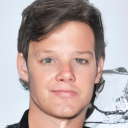} & \includegraphics[width=22mm]{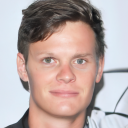} & \includegraphics[width=22mm]{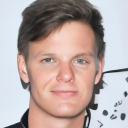} \\
\includegraphics[width=22mm]{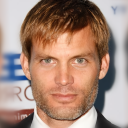} & \includegraphics[width=22mm]{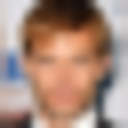} & \includegraphics[width=22mm]{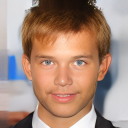} & \includegraphics[width=22mm]{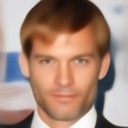} & \includegraphics[width=22mm]{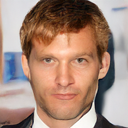} & \includegraphics[width=22mm]{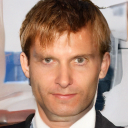} & \includegraphics[width=22mm]{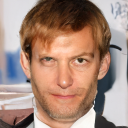} & \includegraphics[width=22mm]{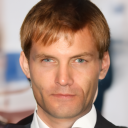} & \includegraphics[width=22mm]{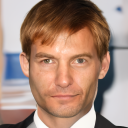} \\
\includegraphics[width=22mm]{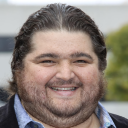} & \includegraphics[width=22mm]{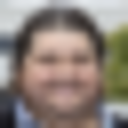} & \includegraphics[width=22mm]{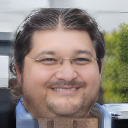} & \includegraphics[width=22mm]{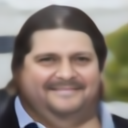} & \includegraphics[width=22mm]{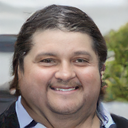} & \includegraphics[width=22mm]{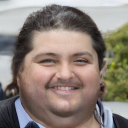} & \includegraphics[width=22mm]{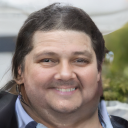} & \includegraphics[width=22mm]{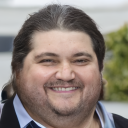} & \includegraphics[width=22mm]{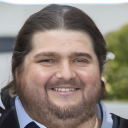} \\
\includegraphics[width=22mm]{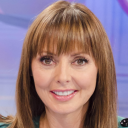} & \includegraphics[width=22mm]{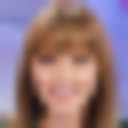} & \includegraphics[width=22mm]{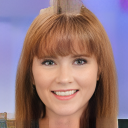} & \includegraphics[width=22mm]{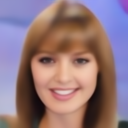} & \includegraphics[width=22mm]{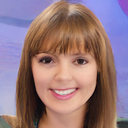} & \includegraphics[width=22mm]{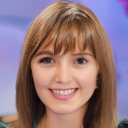} & \includegraphics[width=22mm]{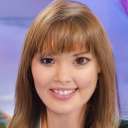} & \includegraphics[width=22mm]{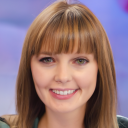} & \includegraphics[width=22mm]{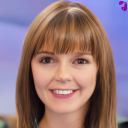}
\end{tabular}
}

\vspace{-1mm}

\caption{Super-resolution results. Our methods are shown in \red{red} (the best) and \blue{blue}. \red{SDE-VE} provides more natural and detailed images than other methods. It is worth highlighting that \blue{\gls*{sde}-\gls*{ve}cs} can even retrieve finer details in some cases, such as the beard in the image shown in the fifth row.}
\label{fig_sr}
\end{figure*}

We evaluate the proposed methods qualitatively (visually) in \cref{fig_sr}, where it is demonstrated that the recovered images with SDE-VE have, in most cases, more quality, more high-frequency details, and are more natural when compared with other methods.
The \sparnet method~\cite{chen2020learning} generates images without distortion, but smoother when compared with  our four methods. For some images, our SDE-VEcs method is able recover even the finest details, such as the beard in the image shown in the fifth row of~\cref{fig_sr}.

\begin{table*}[!ht]
\caption{\gls*{psnr}, \gls*{ssim}, Consistency and CS on $16\times16 \rightarrow 128\times128$ face super-resolution. The best result for CS is highlighted with \red{red}.}
\label{metrics_table}
\centering

\vspace{-2mm}

\begin{tabular}{l c c c c}
\toprule
\textbf{Model} & \textbf{\gls*{psnr}} $\uparrow$ & \textbf{\gls*{ssim}} $\uparrow$ & \textbf{CONSISTENCY} $\downarrow$ & \textbf{CS} $\uparrow$\\
\midrule
\gfpgan~\cite{wang2021gfpgan} & $21.5326\pm 1.5273$ & $0.6006\pm 0.0709$ & $37.2256\pm 12.4622$ & $0.8689\pm 0.0581$\\
\sparnet~\cite{chen2020learning} & $25.2903\pm 1.9770$ & $0.7526\pm 0.0676$ & {$\phantom{0}1.6826\pm \phantom{0}0.7652$} & $0.9371\pm 0.0292$\\
SR3~\cite{saharia2021image} & $22.9581\pm1.8370$ & $0.6605\pm0.0758$ & $\phantom{0}1.3715\pm \phantom{0}0.7904$& $0.9370\pm 0.0244$\\
\gls*{sde}-\gls*{vp} & $22.7171\pm1.8107$ & $0.6448\pm 0.0787$ & $\phantom{0}0.1074\pm \phantom{0}0.0592$ & $0.9330\pm 0.0262$\\
\gls*{sde}-subVP & $22.6455\pm 1.8047$ & $0.6428\pm 0.0797$ & $\phantom{0}0.1433\pm \phantom{0}0.1212$ & $0.9300\pm0.0261$\\
\gls*{sde}-\gls*{ve} & $23.5101\pm 1.9492$ & $0.6879\pm0.0797$ & $\phantom{0}0.0454\pm \phantom{0}0.0357$&\color{red}{$0.9443\pm 0.0222$}\\
\bottomrule
\end{tabular}
\end{table*}

To evaluate our results quantitatively, we present in Table~\ref{metrics_table} the metrics \gls*{psnr}, \gls*{ssim}, consistency and CS.
The \sparnet method~\cite{chen2020learning} achieves the highest value of \gls*{psnr} and \gls*{ssim} \MS{(these values may differ slightly from the original work due to slight differences in the dataset used)}.
However, as discussed earlier, this is not enough to generate images with high-frequency details as it occurs in SR3~\cite{saharia2021image} and in the variations of the proposed approach.
Considering that we intend to apply our super-resolution method to surveillance scenarios and facial recognition, we are interested in recovering details that maximize the CS metric.
As shown in Table \ref{metrics_table}, SDE-VE achieves the highest value for the CS metric, supporting the idea that the SDE-VE method performs better than other super-resolution methods for the face recognition task.
When comparing our four methods based on \glspl*{sde} presented here, the SDE-VE reached the best~results. 

\subsection{Parameters and network training}
\label{subsecd}

Following~\cite{song2021score}, we set the parameters of the models as $\sigma_{min}=0.01$, $\sigma_{max}=348$, $\bar{\beta}_{min}=0.1$, and $\bar{\beta}_{min}=20.0$.
For the optimization, we used the Adam optimizer with a warm-up of $5000$ steps and a learning rate of~$2\times10^{-4}$.

For training, we explored the images from the \ffhq dataset~\cite{karras2019style} with $10^6$ steps. Low-quality images were obtained by downsampling the original high-resolution images by a factor of 8 and upsampling back to the original size using bicubic interpolation.
For testing, considering the high execution time, we used a random sampling of $L=1024$ images of the CelebA-HQ dataset~\cite{karras2018progressive}.
We perform our evaluation in a cross-dataset fashion because it provides a better indication of generalization (hence real-world performance)~\cite{torralba2011unbiased,laroca2022first}.
Previous works also conducted experiments in this way~\cite{chen2020learning,saharia2021image}.
The total number of time steps was fixed in $N=2000$, and for the \gls*{sde}-\gls*{ve}cs, we used $M=2$ correction steps. 
For the VGG-Face feature extraction, a feature vector size of $F=512$ was~used.

All experiments were carried out on a computer with an AMD Ryzen $9$ $5950$X CPU ($3.4$GHz), $128$~GB of RAM ($2633$ MHz), and an NVIDIA Quadro RTX~$8000$ GPU~($48$~GB).

\section{Conclusions and future work}
\label{sec:Conclusions}

In this work, we presented an application in continuous time of diffusion models using \glspl*{sde}. 
The diffusion models have been outperforming \glspl*{gan} for various sets of tasks and can train a model without instabilities (unlike \glspl*{gan}). 
When using facial features as metrics combined with qualitative analysis, we demonstrated that the \gls*{sde}-\gls*{ve} model reaches better results than other methods for the super-resolution~task. 

The \gls*{sde}-\gls*{ve} super-resolution algorithm also has excellent potential to be used for recognition tasks.
\MS{
As a next step, we intend to demonstrate the effectiveness of our method on images from surveillance camera datasets such as QUIS-CAMPI~\cite{neves2018quis} and SCface~\cite{grgic2011scface}.
}

The influence of the image smoothness and \gls*{psnr} values on the recognition task will be further explored in future works. More specifically, explainability techniques can be used to analyze which characteristics change when we alter the image smoothness and \gls*{psnr} values,  similar to the work performed for periocular recognition in~\cite{brito2021short}.

Despite the superior results of general diffusion models and the method presented in this work, which is based on \glspl*{sde}, we remark that all these methods \MS{have a relatively high  execution time}.
Thus, it is important to research novel strategies to improve the efficiency of these methods.

\balance
\section*{\uppercase{Acknowledgments}}

\iffinal
    This work was partly supported by the Coordination for the Improvement of Higher Education Personnel~(CAPES) (\textit{Programa de Coopera\c{c}\~{a}o Acad\^{e}mica em Seguran\c{c}a P\'{u}blica e Ci\^{e}ncias Forenses \#~88881.516265/2020-01}), and partly by the National Council for Scientific and Technological Development~(CNPq) (\#~308879/2020-1).
    We gratefully acknowledge the support of NVIDIA Corporation with the donation of the Quadro RTX $8000$ GPU used for this~research.    
\else
    The acknowledgments are hidden for review.
\fi

\bibliographystyle{IEEEtran}
\bibliography{bibtex}

\end{document}